\pdfoutput=1

\documentclass[11pt]{article}

\usepackage[final]{EMNLP2023}

\usepackage{times}
\usepackage{latexsym}

\usepackage[T1]{fontenc}

\usepackage[utf8]{inputenc}

\usepackage{microtype}

\usepackage{inconsolata}

\usepackage{url}
\usepackage{hyperref}
\usepackage{array}
\usepackage{pifont}
\usepackage{tabularx}
\usepackage{adjustbox}
\usepackage{multirow}
\usepackage{enumitem}
\usepackage{xspace}
\usepackage{tcolorbox}
\usepackage{subcaption}
\usepackage{booktabs,amsfonts,dcolumn}
\usepackage{amsmath,amsthm,amsfonts,amssymb,bm,stmaryrd,bbm}
\usepackage{makecell}
\usepackage{graphicx}
\usepackage{colortbl}
\usepackage{color,soul}
\usepackage{float}
\usepackage[colorinlistoftodos]{todonotes}

\title{Towards Practical Tool Usage for Continually Learning LLMs}

\author{
    Jerry Huang \\ Mila - Quebec AI Institute\\ Universit\'{e} de Montr\'{e}al \\ {\rm \texttt{\href{mailto:jerry.huang@mila.quebec}{jerry.huang@mila.quebec}}}
    \AND Prasanna Parthasarathi \\ Huawei Noah's Ark Lab
    \And Mehdi Rezagholizadeh \\ Huawei Noah's Ark Lab
    \And Sarath Chandar \\ Mila - Quebec AI Institute \\ Polytechnique Montr\'{e}al \\ Canada CIFAR AI 
}

\begin{document}

\def\chapterautorefname{Chapter}%
\def\sectionautorefname{Section}%
\def\subsectionautorefname{Subsection}%
\def\subsubsectionautorefname{Subsubsection}%
\def\paragraphautorefname{Paragraph}%
\def\tableautorefname{Table}%
\def\equationautorefname{Equation}%

\providecommand{\draft}[1]{\colorbox{green}{\color{cyan}{#1}}}
    
\providecommand{\todoall}[1]{
    {\protect\color{red}{[TODO: {#1}]}}
}

\providecommand{\jerry}[1]{
    {\protect\color{purple}{[Jerry: {#1}]}}
}

\providecommand{\prasanna}[1]{
    {\protect\color{orange}{[Prasanna: {#1}]}}
}

\providecommand{\sarath}[1]{
    {\protect\color{blue}{[Sarath: {#1}]}}
}

\providecommand{\mehdi}[1]{
    {\protect\color{cyan}{[Mehdi: {#1}]}}
}

\newcolumntype{P}[1]{>{\centering\arraybackslash}p{#1}}
\newcommand{\mydots}{\hbox to 1em{.\hss.\hss.}}

\newcommand{\cmark}{\ding{51}}
\newcommand{\xmark}{\ding{55}}

\newcommand\sys[1]{\textsc{#1}}
\newcommand\ti[1]{\textit{#1}}
\newcommand\ts[1]{\textsc{#1}}
\newcommand\tf[1]{\textbf{#1}}
\newcommand\ttt[1]{\texttt{#1}}
\newcommand\mf[1]{\mathbf{#1}}
\newcommand\tmp[1]{\color{gray}{#1}}
\newcommand\warn[1]{\textbf{\color{red}{#1}}}
\newcommand\mr[1]{\mathrm{#1}}
\newcommand\mc[1]{\mathcal{#1}}

\newcommand{\cls}{\ttt{[CLS]}}
\newcommand{\mcorr}{m_{\text{corr}}}
\newcommand{\mpred}{m_{\text{pred}}}

\newcommand\myeq{\stackrel{\mathclap{\normalfont\mbox{i.i.d.}}}{~}}
\providecommand{\todon}{
    {\protect\color{red}{00.00}}
}

\newcommand{\uours}{USimCSE\xspace}
\newcommand{\ours}{SimCSE\xspace}
\newcommand{\identical}{identical}
\newcommand{\Identical}{Identical}

\newcommand{\la}{$_\texttt{large}$}
\newcommand{\ba}{$_\texttt{base}$}

\renewcommand{\paragraph}[1]{\vspace{0.2cm}\noindent\textbf{#1}}
\newcommand{\tpf}[1]{\noindent\textbf{#1}}
\newcommand{\tableindent}{~~}

\definecolor{ccon}{HTML}{fee9d4}
\definecolor{cood}{HTML}{d8f0d3}
\definecolor{cid}{HTML}{dae8f5}

\definecolor{gred}{HTML}{cc0200}
\definecolor{ggreen}{HTML}{38761c}

\newcommand{\up}{\textcolor{ggreen}{$\uparrow$}}
\newcommand{\down}{\textcolor{gred}{$\downarrow$}}
\newcommand{\neu}{\textcolor{cid}{$\downarrow$}}
\newcommand{\neuwhite}{\textcolor{white}{$\downarrow$}}
\newcommand{\recipe}{efficient pre-training recipe}
\newcommand{\academicbert}{24hBERT}
\newcommand{\eight}{\mbox{80-10-10}}
\newcommand{\corrx}{\tilde x}

\definecolor{c1}{cmyk}{0,0.6175,0.8848,0.1490}
\definecolor{c2}{cmyk}{0.1127,0.6690,0,0.4431}
\definecolor{c3}{cmyk}{0.3081,0,0.7209,0.3255}
\definecolor{c4}{cmyk}{0.6765,0.2017,0,0.0667}
\definecolor{c5}{cmyk}{0,0.8765,0.7099,0.3647}

\newtcbox{\hlprimarytab}{on line, rounded corners, box align=base, colback=c3!30,colframe=white,size=fbox,arc=3pt, before upper=\strut, top=-2pt, bottom=-4pt, left=-2pt, right=-2pt, boxrule=0pt}
\newtcbox{\hlsecondarytab}{on line, box align=base, colback=red!30,colframe=white,size=fbox,arc=3pt, before upper=\strut, top=-2pt, bottom=-4pt, left=-2pt, right=-2pt, boxrule=0pt}

\newcommand{\dashifted}{\raisebox{0.5\depth}{\tiny$\downarrow$}}
\newcommand{\uashifted}{\raisebox{0.5\depth}{\tiny$\uparrow$}}
\newcommand{\da}[1]{{\small\hlsecondarytab{\dashifted{#1}}}}
\newcommand{\ua}[1]{{\small\hlprimarytab{\uashifted{#1}}}}

\newcommand{\db}[1]{{\small\hlprimarytab{\dashifted{#1}}}}
\newcommand{\ub}[1]{{\small\hlsecondarytab{\uashifted{#1}}}}

\newcommand{\papersummary}[3]{
    \begin{itemize}
        \item {\bf What}: #1
        \item {\bf How}: #2
        \item {\bf Difference}: #3
    \end{itemize}
}

\makeatletter
\def\thickhline{%
  \noalign{\ifnum0=`}\fi\hrule \@height \thickarrayrulewidth \futurelet
   \reserved@a\@xthickhline}
\def\@xthickhline{\ifx\reserved@a\thickhline
               \vskip\doublerulesep
               \vskip-\thickarrayrulewidth
             \fi
      \ifnum0=`{\fi}}
\makeatother
\newlength{\thickarrayrulewidth}
\setlength{\thickarrayrulewidth}{2\arrayrulewidth}

\maketitle

\begin{abstract}
Large language models (LLMs) show an innate skill for solving language based tasks. But insights have suggested an inability to adjust for information or task-solving skills becoming outdated, as their knowledge, stored directly within their parameters, remains static in time. Tool use helps by offloading work to systems that the LLM can access through an interface, but LLMs that use them still must adapt to nonstationary environments for prolonged use, as new tools can emerge and existing tools can change. Nevertheless, tools require less specialized knowledge, therefore we hypothesize they are better suited for continual learning (CL) as they rely less on parametric memory for solving tasks and instead focus on learning when to apply pre-defined tools. To verify this, we develop a synthetic benchmark and follow this by aggregating existing NLP tasks to form a more realistic testing scenario. While we demonstrate scaling model size is not a solution, regardless of tool usage, continual learning techniques can enable tool LLMs to both adapt faster while forgetting less, highlighting their potential as continual learners.
\end{abstract}

\section{Introduction}

Performance of pre-trained LLMs~\citep{raffel2020exploring, chung2022scaling, touvron2023llama} on a variety of domains~\citep{srivastava2023beyond, OpenAI2023GPT4TR}, and probing the parameters~\citep{petroni-etal-2021-kilt} validate that LLMs possess a representation of knowledge in their parameters. 
However, such knowledge across domains expires at differential rates---\emph{What is the current population of USA?} becomes obsolete in a decade while \emph{Who is the President of X} expires in expectation around every $4$ years, and say \emph{What is the current interest rate?} expires more frequently. This affects model performance largely because that these models store information directly as \textit{parametric knowledge}~\citep{petroni-etal-2019-language} and retrieve them when prompted~\citep{parameters}. Alternatively, even if the information within the world does not change at once, the world may change in such a way that the goal of the LLM changes~\citep{kenton2021alignment}. Hence the consensus is that the generated responses from pre-trained LLMs become unreliable~\citep{zhang-choi-2021-situatedqa, komeili-etal-2022-internet} and the LLMs have to adapt to make its generated texts relevant. 


The vanilla approach to avoid staleness is to collect more data that better reflects the current world and re-train from scratch~\citep{gao2020pile}. The disadvantage is that the necessary resources grow with the data and since models store information directly within parameters, additional parameters are needed to hold the new knowledge~\citep{jang2022towards}.
Two popular alternative solutions are pursued:
One---\textit{knowledge editing}~\citep{de-cao-etal-2021-editing}--- is based on the assumption that knowledge in LLMs' parameters can be updated by modifying the parameters directly. But editing factual knowledge can warp the innate knowledge structure of LLMs~\citep{gupta-etal-2023-editing} and approaches that do not directly intervene on the parameters require the use of additional memory~\citep{mitchell22memory, dong-etal-2022-calibrating}. Another is the usage of low-rank adapters~\citep{hu2022lora}, which freezes a base model and introduces smaller \textit{adapters} which can be used to fine-tune the model for down-stream tasks without needing to train it explicitly. However, adapters are task specific, meaning this can be costly once the number of tasks has grown, and it is the adapter that is tasked with handling changes in the data rather than the model itself.

Tangential to the knowledge forgetting problem, LLMs are trained to use tools~\citep{schick2023toolformer} through APIs and retrieve information from outside sources rather than parameters directly~\citep{rag}. 
Furthermore, with tool API the information being stored outside of LLMs allow for independent updates and a model using it only requires maintaining updates to the tools usage to remain up-to-date. Though this provides a reasonable simplification to the differential expiry rates in knowledge, tool-use itself does not make LLMs everlasting, as both the tools themselves and the set of existing tools can change, which tools LLMs must adapt to. As such, tool-use itself is insufficient for the non-stationary setups as discussed in the \textbf{continual learning (CL)} literature~\citep{ring1998child, thrun1998lifelong}, where it is the model that must learn to autonomously adapt to change in either the state of the world as well as down-stream tasks. Within this setting, this points at the non-stationarity in the tool definition 
which can inherently lead to difficulties adjusting to distribution shifts, as learned features for specific tasks often cannot adapt to new ones~\citep{distort-features}. 

Such simplification of complex tasks 
also runs the risk of overfitting to present tasks, leading to forgetting the past~\citep{catastrophic-forgetting, forgetting-2, forgetting} by large parameteric models. A careful treatment is therefore needed to modify the static knowledge repository of LLMs into models capable of continually adapting to the non-stationarity involved in learning tools that vary in complexity. 
We summarize our work as follows:
\begin{enumerate}
  \item We propose a synthetic arithmetic dataset with Easy and Difficult splits, and benchmark LLMs of size 125M-13B on using the tools in a task of continual API learning.
  \item We show that even with scale, LLMs are incapable of naively adapting to task shifts through sequential fine-tuning highlighting the drawback of mere parametric knowledge to handle
  distribution shifts.
  \item However, with a replay buffer,
  we demonstrate that tool LLMs can adapt to these task shifts, whereas standard LLMs still fall short.
\end{enumerate}

\section{Related Works}

\paragraph{LLMs as Continual Learners.} 
Learning in a non-stationary setting has been treated formally in the continual learning~\citep{chen2018lifelong} (CL) paradigm. The objective of CL~\citep{thrun1998lifelong, kirkpatrick2017overcoming} is to learn from a sequence of tasks without the 
forgetting~\citep{forgetting-2} of previously seen tasks. With growing emphasis on language based applications, CL in training of LLMs has focused on two main directions:
\begin{enumerate}[label=(\arabic*)]
    \item Task learning, where LLMs must learn multiple downstream tasks in sequence~\citep{huang-etal-2021-continual, mehta2022an}.
    \item Domain adaptation, where the LLM is trained on multiple data domains~\citep{gururangan-etal-2020-dont, ke2023continual} and must remain knowledgeable about each.
\end{enumerate}
However, LLMs with large parameteric spaces limit the applicability of regularization-based techniques~\citep{li2017learning, lopez2017gradient, zenke2017continual, aljundi2018memory} while the few-shot abilities of LLMs~\citep{brown2020language} suggest accommodating replay buffers~\citep{rebuffi2017icarl, lopez2017gradient, shin2017continual, chaudhry2018efficient, wang2019sentence} of intractable sizes. 

\paragraph{Efficiently Updating LLMs.} 
Because LLMs are so costly to train~\citep{strubell-etal-2019-energy}, delaying their expiry date requires being able to update knowledge cheaply~\citep{zhang2024comprehensive}. 
Within this space, two types of methods, parameter-preserving and parameter-editing, have emerged. Parameter-preserving methods, focus on keeping the underlying model intact~\citep{dong-etal-2022-calibrating, huang2023transformerpatcher, hartvigsen2023aging, zhong-etal-2023-mquake}. 
Additional parameters or memory to track stale facts could quickly become impractical as the number of edits increases. 
Alternatively, parameter-editing methods directly modify the model parameters through fine-tuning the model to update only a select set of parameters~\citep{zhu2021modifying, lee-etal-2022-plug}, meta-learning the parameters to edit~\citep{mitchell2022fast}, or locating and modifying the relevant parameters~\citep{santurkar2021editing, tanno2022repairing}. This results in fast edits with little to no memory overhead. Yet the complicated structure of LLMs makes this a risky proposition, as modifying even one parameter can have various unknown downstream effects that can affect the usability of the model~\citep{chen2023journey}.


\paragraph{Tool-Augmented LLMs.} LLMs are generalist agents that can be adapted to perform on a wide range of natural language tasks~\citep{brown2020language, chowdhery2022palm}. 
However, they still struggle in specialized settings~\citep{patel-etal-2021-nlp, lin-etal-2022-shot}
and have issues disassociating entities from extra-linguistic~\citep{zhang-choi-2021-situatedqa} or even spurious~\citep{joshi-etal-2022-spurious} contexts. 

Tool-augmented LLMs~\citep{schick2023toolformer} address this by learning to manipulate \textit{specialized tools} to handle the knowledge-based computations. \citet{wang2022code4struct, imani2023mathprompter, paranjape2023art} have shown improved zero-shot performance across a variety of downstream tasks without drops in language modeling abilities. Tools simplify tasks for LLMs, 
potentially reducing solving a task to learning to route to appropriate tools. However, these prior works do not study how tool LLMs adapt to new tasks or settings. 

This work attempts to measure the issues that stem from LLMs forgetting by directly learning sequentially through the task samples. By replacing direct-learning with learning with tools,
the work reposes the tasks in the tool space, and solves a unified non-stationarity problem of continual learning of tools as a proxy to solve the challenge of continual learning using task samples directly.


\begin{figure*}[ht]
    \centering
    \includegraphics[width=0.95\linewidth]{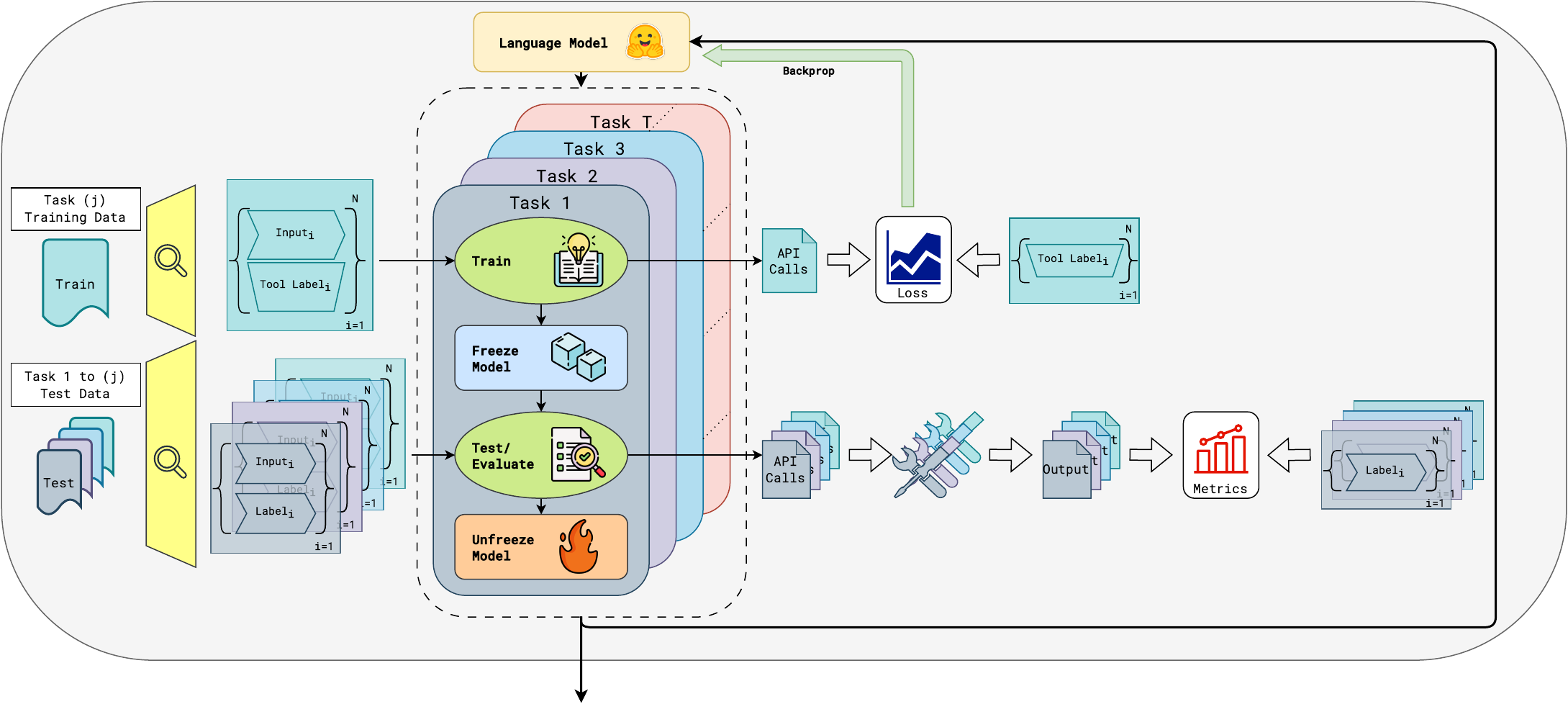}
    \caption{CL with Tools - For a task, the model is first trained to predict/generate tool calls, rather than explicit responses. The trained model is then frozen and evaluated, during which it outputs tool calls that are parsed and executed to return an output which is compared against the ground-truth. The model is then unfrozen and trained on the next task in the sequence. This is repeated until all tasks have been learned by the model.}
    \label{fig:setup}
\end{figure*}

\section{Motivating Questions}
\label{subsec:motivating-questions}

More formally, continually adapting LLMs to the changing world and domain knowledge is a complex but relevant problem, as forgetting prior information can limit the applicability of LLMs. Further, with shifts in domain being aperiodic for diverse knowledge and LLMs being the generalist model they are leads us to the pertinent question:

\begin{quote}
    \textbf{\textit{Can learning to use tools alleviate sequential learning challenges?}}
\end{quote}
and the sub-questions that need to be answered:
\begin{enumerate}[label=(\textsc{Q\arabic*})]
    \item \textit{How far can we push by simply increasing parametric knowledge space help for continual learning?}
    \item \textit{Are there limits to how much both tool LLMs and vanilla LLMs can learn continually?}
    \item \textit{How do tool LLMs fare with imperfect tools?}
\end{enumerate}
We use these questions to build our methodology and experimental design in the following sections.

\section{Methodology}
\label{sec:methodology}

\subsection{Preliminaries}

\paragraph{Model:} We use causal Transformer-based language models in a text-generation setup, in particular, the OPT~\citep{zhang2022opt} family of pre-trained LLMs up to $13$B parameters. This allows us to compare the powerful similar generative language models with scale.


\paragraph{Dataset Format:} To properly assess the usefulness of tool learning, each sample $s \in \mathcal{D}$ consists of a query $s_q$, the raw answer to the query, $s_G$, and an API call answer $s_A$, which can be executed by a task-specific API to obtain a response that is compared with $s_G$ using exact string matching.

\paragraph{Learning Setup:} Language models are trained either \textbf{with tools} or \textbf{without tools} to solve a sequence of $T$ tasks---$\{\mathcal{T}_1, \dots, \mathcal{T}_T\}$. Each task $\mathcal{T}_k$ defines a specific tool and a dataset $\mathcal{D}_k$ which contains the examples associated with learning the $k$-th tool.   
{\bf With Tools} the model learns to generate the API calls, as mentioned previously, that gets routed to appropriate API to generate the answer. {\bf Without tools}, the model is fine-tuned to predict the answer directly, such as a numerical or textual response.
Iterating over tasks in sequence, at every iteration,$i$, a model is trained with examples corresponding to $\mathcal{T}_i$ and evaluated on test sets of all the tasks the model has seen until then. Each task uses a learning rate warm-up followed by a decay to $0$, \emph{i.e.} the learning rate warm-up and decay repeats for each task in the set. We use the AdamW~\citep{loshchilov2018decoupled} optimizer with a peak learning rate based on the model size\footnote{Hyper-parameters are provided in~\autoref{app:hyperparameters}}.




\subsection{Baselines}
\label{subsec:baselines}
For each setup, we train under a number of settings:

\paragraph{Sequential Fine-tuning:} The model sees a stream of tasks $\{\mathcal{T}_i\}_{i=1}^T$  in an order without repetition. The model is explicitly fine-tuned on each task and once complete moves to the next task for training. 

\paragraph{Mixed Dataset:} All tasks are mixed into a single task to train a model. This is equivalent to ``seeing'' all tasks at once and is a strong upper bound, where model learns from all available data at once.


\paragraph{Episodic Replay (ER):} \citet{chaudhry2019tiny} augment models with a replay buffer that retains examples from the previous tasks. With the buffer, the model continually takes some of the recent data and randomly replaces older samples. When training, the model will randomly sample a batch from the replay buffer and calculate a replay loss which is added to the standard loss before performing a gradient update. Motivating the usage of this method are observations that LLMs are few-shot learners~\citep{brown2020language}, suggesting that this may be an efficient use case of the method given the smaller number of examples and subsequent buffer size that may be necessary.

\subsection{Evaluation Metrics}
\label{subsec:evaluation}


We evaluate using the following metrics for measuring performance\footnote{Formulas and further details are provided in \autoref{app:eval}.}:

\paragraph{Accuracy}: Given that each task $\{\mathcal{T}_1, \dots, \mathcal{T}_T\}$ consists of a train and test set, we can measure the accuracy on each test set individually. 
We report the average accuracy on test sets up to the most recent task on which the model was trained.  
In particular, suppose a model is being trained on task $\mathcal{T}_\tau$. The average accuracy is measured as in \autoref{eqn:acc},

\begin{equation}
    \hat{p}_\tau = \frac{1}{\tau}\sum^{\tau}_{k=1}p_{\tau,k}
    \label{eqn:acc}
\end{equation}
where $p_{\tau, k}$ denotes the performance of the model on the test set associated with task $k$ after having trained on task $\tau$.
In the tool setup, $p_{\tau,k}$ is measured by parsing and executing the generated API calls and computed exact match~\citep{rajpurkar-etal-2016-squad} with the true answers.

\paragraph{Forgetting}~\citep{chaudhry2018riemannian}: 
Forgetting is the average degradation ($\geq 0$) in performance on all seen tasks excluding the most recent task on which the model was trained. 


\begin{equation}
    f_{\tau} = \frac{1}{\tau-1}  \sum_{k=1}^{\tau-1} \underset{t \in \{1,\dots \tau-1\}}{\text{max}}\bigg(\frac{p_{t, k} - p_{\tau, k}}{p_{t, k}}, 0\bigg),
    \label{eqn:forgetting}
\end{equation}
\autoref{eqn:forgetting} measures the average forgetting a model has after trained on $\mathcal{T}_\tau$.

\paragraph{Learning Accuracy}~\citep{riemer2018learning}: Models have limited capacity and can learn a limited number of tasks. 
\begin{equation}
L_{\tau} = \frac{1}{\tau} \sum_{k=1}^\tau p_{k,k},
    \label{eqn:learning_accuracy}
\end{equation}
\autoref{eqn:learning_accuracy} defines learning accuracy (L-A) to approximate the learning capacity by measuring average performance on each task immediately after being trained on it.
\section{Experiments}
\label{sec:experiments}

We design our experiments to address \textsc{Q1-3} in \S \ref{subsec:motivating-questions} as follows.
\begin{figure*}[!t]
    \centering
    \begin{subfigure}[b]{0.3\textwidth}
        \includegraphics[width=\textwidth]{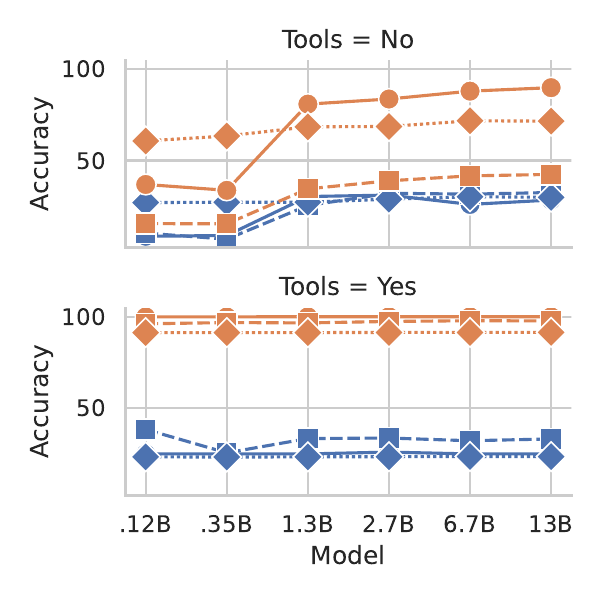}
        \caption{ }
        \label{fig:Accuracy-all-tasks}
    \end{subfigure}%
    \hfill
    \begin{subfigure}[b]{0.3\textwidth}
        \includegraphics[width=\textwidth]{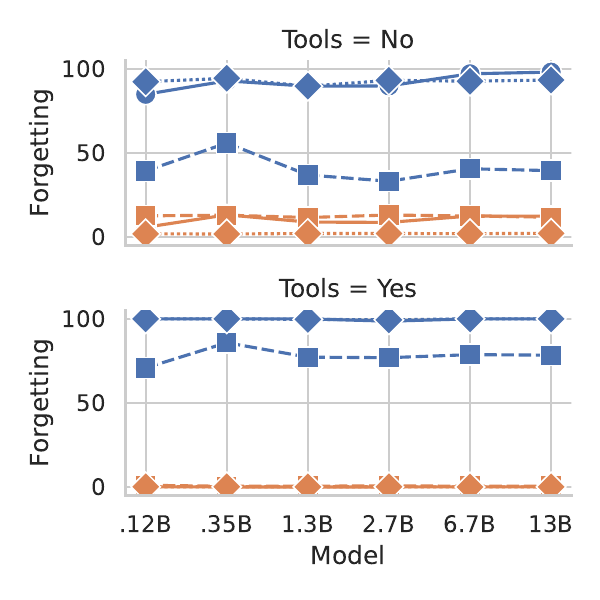}
        \caption{ }
        \label{fig:Forgetting-all-tasks}
    \end{subfigure}
    \hfill
    \begin{subfigure}[b]{0.38\textwidth}
        \includegraphics[width=\textwidth]{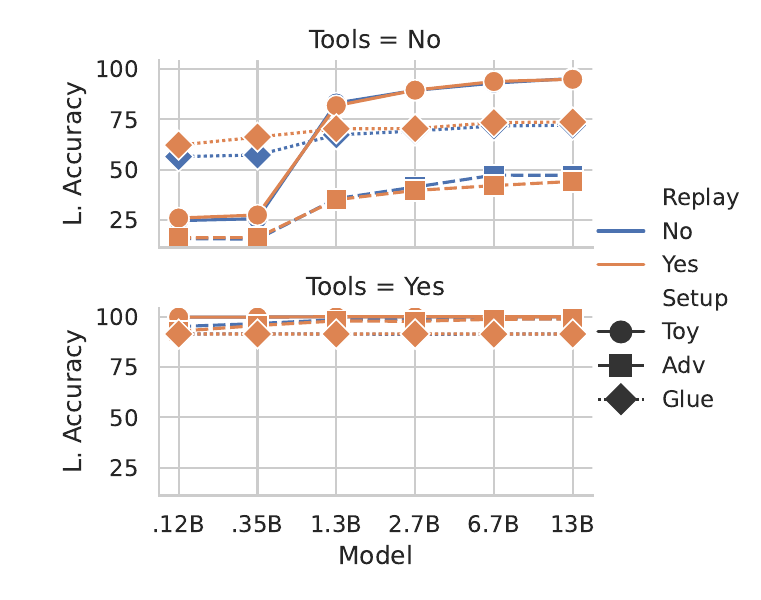}
        \caption{ }
        \label{fig:LAcc-all-tasks}
    \end{subfigure}%
    \caption{ Across the different task setups, we measure the different metrics. Although it is evident that using tools improves the L-Accuracy significantly, we observe that the Accuracy across tasks is not reflecting the same. The significant forgetting of tools only get fixed with appropriate usage of a replay buffer to improve the overall accuracy irrespective of the task difficulty.}
    \label{fig:master-results}
\end{figure*}

\paragraph{Does more parametric space solve CL?} Understanding the effects of scale requires ensuring that larger models can adequately solve the task when not presented in a sequential learning setting. 

For \textsc{Q1}, we first construct a synthetic arithmetic dataset consisting of four functions, each with a single template and limited to integers between $0$ to $99$. Every sample in the set adheres to the format defined in \S \ref{sec:methodology}. Each operation has an associated API format answer--- ex. \textsc{Add($a$, $b$)} where $a$ and $b$ are arguments provided to the tool--- and explicit numerical answers. Non-integers are expressed in decimal format, \emph{e.g.} $0.75$. Tasks have the same number of examples, divided into training and test sets\footnote{Additional hyper-parameters are found in \autoref{app:hyperparameters}}. We refer to this as our \emph{Toy Arithmetic Task}. 

Using this task, we verify if increasing the number of parameters of LLM improves performance, measured through with accuracy and forgetting. If accuracy increases while forgetting decreases with the parameters, then we can infer that link between the size of the parametric space and performance is in fact linked. However, we observe in \S\ref{sec:results} that this is far from the case.

\paragraph{How much can LLMs learn continually?} Following the observations from our previous question, we further the study to find the extent LLMs can learn continually when the task gets difficult. 
As such, the next goal is to observe some of these limits in both with and without using tools by increasing the task difficulty, for which we build a more difficult arithmetic benchmark. 

We add difficulty through expanding input and output space along with adding ambiguous templates. To expand the input space, we create additional functions and templates for existing functions which must be learned to properly use the tools. Furthermore, we add more ambiguity to the tool templates by increasing the token similarity for many templates. The output space now covers operands which can include real numbers, hence requiring the models to be capable of identifying which parts of the input properly constitute parts of the tools calls and use this to create the output. We refer to this as our \emph{Advanced Arithmetic Task}.

The goal here is to observe if performance (as measured in the same way in the previous task) can remain constant between the two tasks both with and without using the tools. We expect that although accuracy may drop in part due to the task difficulty, but does forgetting remain consistent between the toy task and this more difficult one? In particular, if forgetting is consistently greater for the advanced task, this highlights limitations in the LLM with respect to becoming more general, autonomous multi-task learners.





\begin{table}[ht]
    \centering
    \resizebox{\linewidth}{!}{%
    \begin{tabular}{lll}
        \toprule
        \textbf{Description} & \textbf{API Format} & \textbf{Example} \\
        \midrule
        Add two numbers & \textsc{add(a, b)} & $\textsc{add(23, 35)} = 58$ \\
        Subtract $b$ from $a$ & \textsc{sub(a, b)} & $\textsc{sub(34, 12)} = 12$ \\
        Multiply two numbers & \textsc{mult(a, b)} & $\textsc{mult(5, 7)} = 35$ \\
        Divide $a$ by $b$ & \textsc{div(a, b)} & $\textsc{div(81, 2)} = 41.5$ \\
        \midrule
        Compute the GCD & \textsc{gcd(a, b)} & $\textsc{gcd(12, 20)} = 4$ \\
        Compute the LCM & \textsc{lcm(a, b)} & $\textsc{lcm(14, 21)} = 42$ \\
        List the prime factors & \textsc{lp(a)} & $\textsc{lp(4) = 1, 2}$\\
        \bottomrule
    \end{tabular}}
    \caption{Tasks within the toy setup (top) and additional operations in the advanced setup (bottom).}
    \label{tab:operations}
\end{table}

\paragraph{Do tool LLMs behave well with imperfect tools?} While our arithmetic benchmarks are useful for evaluating LLMs in a continual learning setting, the goal remains to use them for more practical use cases. In such cases, oracles are not readily available in the same way as in the arithmetic settings. Explicit rules exist for arithmetic tasks; if one designs a tool to follow said rules, then mastering the tool is equivalent to mastering the task. But this assumption fails to hold in many cases, which motivates our next question: do the benefits of tool learning still hold once we move away from perfect tools.
For this, we construct a more representative benchmark where the LLM attempts to learn tools that correspond to different natural language understanding tasks. 

We use a subset of tasks from the GLUE benchmark~\citep{wang2018glue}, in particular MNLI, QQP, SST-2 and CoLA, to be learnt continually.
Each task assumes a specific tool. For example, samples from QQP ask if two questions are paraphrases of each other; we provide a tool response with a tool \textsc{paraphrase} requiring two string arguments. For the tool response, we use the API answer as input to a separate language model which returns the final response as the tools are no longer oracles; they can produce an incorrect answer.

This aims to address \textsc{Q3} by providing a more realistic scenario for continual learning where the tools themselves cannot completely solve the task. Here, we wish to verify if the gaps we observe in the arithmetic setups disappear simply through the use of imperfect tools, again through the lens of accuracy and forgetting.
\section{Results and Analysis}
\label{sec:results}




\paragraph{LLMs struggle with CL, irrespective of tools.} 
In \autoref{fig:master-results}, we compare the performances of the different sized architectures on the synthetic arithmetic datasets, and the realistic task as described in \S\ref{sec:experiments}.
As we experiment both directly learning over the samples and learning to use APIs, we observe that generalizing on arithmetic tasks is challenging to learn directly from samples (Tools=No in \autoref{fig:master-results}). Also the forgetting (\autoref{fig:Forgetting-all-tasks}) is significant irrespective of the models using tools to solve the task. Though the learning accuracy for even smaller sized LMs was higher with tools as compared to $100\times$ larger model without using tools, we observe  the retention of past tasks as observed in Accuracy in \autoref{fig:Accuracy-all-tasks} appears as a prevalent issue across the model sizes.
\begin{figure}[ht]
    \centering
        \includegraphics[width=\linewidth]{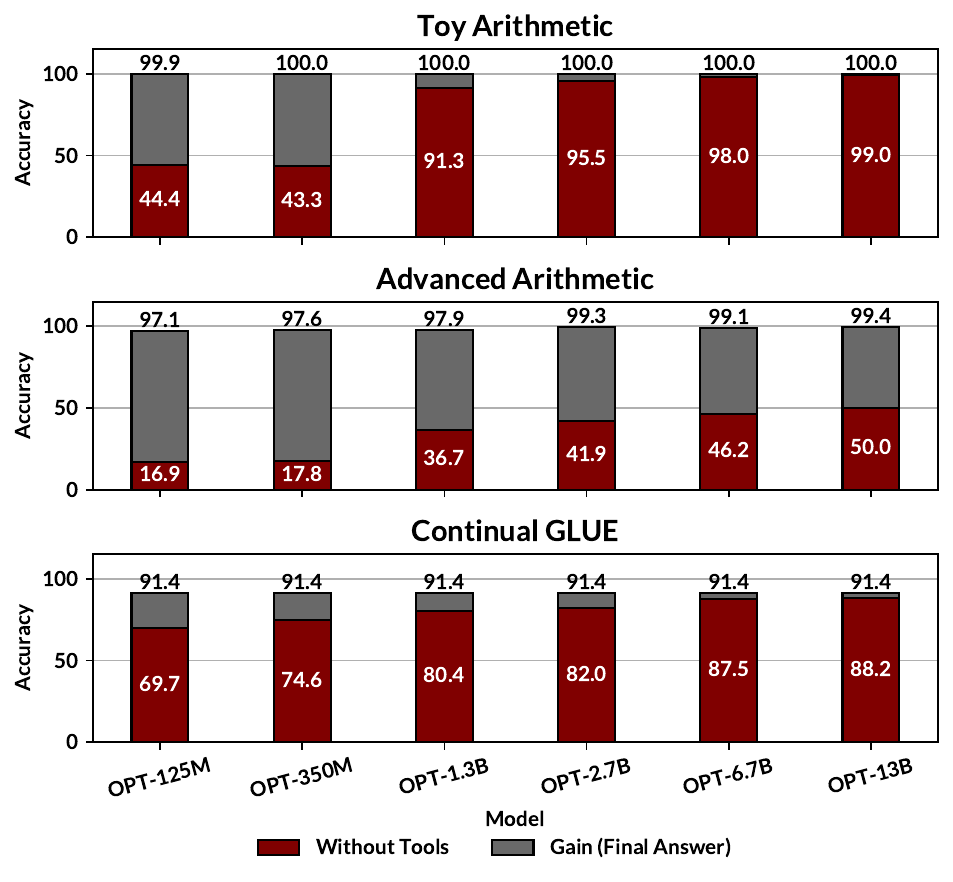}
        \caption{Accuracy on all benchmarks when tasks are mixed (5 seeds). Red bars note accuracy without tools, grey bars show the gain from using tools. Top labels show accuracy using tools. Tabular versions of numerical results are available in \autoref{app:details}.}
        \label{fig:mixed-figure}
\end{figure}

While the results demonstrate the effect of LLMs struggle with sequential learning, we look at whether the performance degradation is an artifact that comes with the learning set up. To that, we compare the performances of the models in a mixed dataset setting where the models learn all the tasks at once with and without using the tools. The hypothesis is that if the LMs showed significant retention as indicated with the comparable performances to using tools, it can be regarded that more data potentially solves the forgetting problem. But, to the contrary in \autoref{fig:mixed-figure} we observe that the gap does exist in the different tasks. So, irrespective of using tools or task seen all at once or not LLMs struggle with the generalizing to the tasks.

\begin{figure}[ht]
    \centering
    \begin{subfigure}[b]{0.45\textwidth}
        \includegraphics[width=\textwidth]{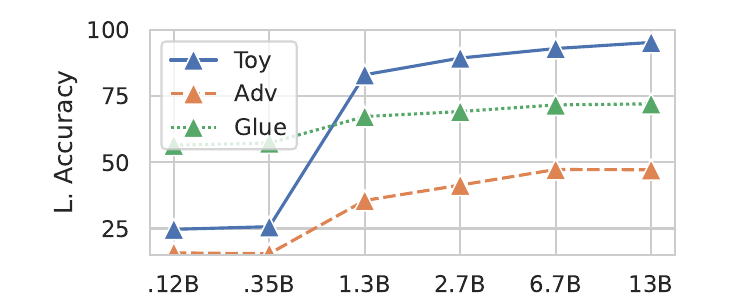}
        \caption{}
        \label{fig:LAccuracy_with_size}
    \end{subfigure}
    
    \vspace{\floatsep}  
    
    \begin{subfigure}[b]{0.45\textwidth}
        \includegraphics[width=\textwidth]{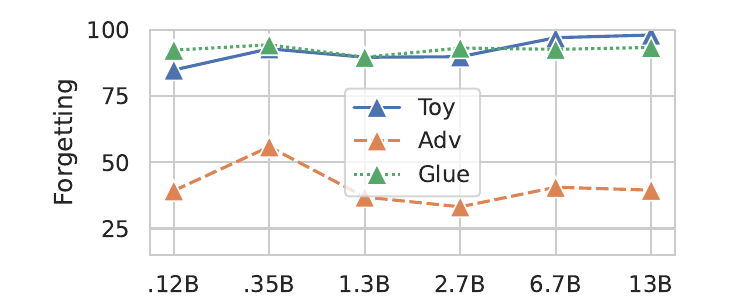}
        \caption{}
        \label{fig:Forgetting_with_size}
    \end{subfigure}
    
    \caption{While we observe that the scale (when not using tools) plays a significant role in how the model's capacity is used to learn a task, the lack of similar effect with forgetting suggests a 13B model is only as good as a 125M model in retaining knowledge of past tasks.}
    \label{fig:twosubfigures}
\end{figure}

\paragraph{More parameters $\neq$ Less forgetting.} \autoref{fig:LAccuracy_with_size} indicate the effect of model size on the ability of learning tasks to increase with model size. However, from \autoref{fig:Forgetting_with_size}, we fail to see any systematic decrease in the forgetting of the model, suggesting that being able to learn tasks sequentially remains a concern despite the increase in model capacity. Nevertheless, the greater learning accuracy observed with larger models can be useful to unleash the potential of tool LLMs.

In particular, we observe in \autoref{fig:LAcc-all-tasks} that tool LLMs' learning accuracy to be consistently higher than vanilla LLMs, suggesting a faster adaptation with tools. Even more encouraging is the fact that learning accuracy for the smallest tool LLMs is often far superior compared to the largest vanilla LLMs. This is promising, as it demonstrates that if one can overcome the forgetting concern that plagues LLMs in general, then using tool LLMs may be much more efficient than vanilla LLMs as they can replace ones that are larger for similar performance. This observation not only is evident when the tools are non-parametric oracles as in our arithmetic tasks but also in the case of our continual GLUE task where tools themselves are parametric models. Though models are no longer oracles, as demonstrated by imperfect learning accuracy (\autoref{fig:LAcc-all-tasks}), the combined parametric space with smaller experts is still significantly smaller than a vanilla LLM that achieves equivalent performance.

By reposing problems in the tool space, models learn only to make the correct API calls and we see smaller models with tools to perform on par with larger models not using tools. Beyond a simplistic comparison, this could also be seen as an economic way to guarantee consistency and truthfulness to the results while not incurring the cost of pre-training larger LLMs as the reliance is less on the more riskier LLMs' parametric knowledge~\citep{kazemnejad2023measuring}.

These results motivate potential opportunities in building smaller models and learnable API calls that can outsmart large LLMs in terms of efficiency with cheaper training costs. While LLMs trained for more complex interaction and usage exist, such as instruction fine-tuned ones~\citep{askell2021general, ouyang2022training, dubois2023alpacafarm}, they still rely on the assumption that the underlying world does not change; one can still expect false statements unless they are explicitly trained to rely on outside data sources accessible in a predetermined manner. As such, tool LLMs present an opportunity to move away from larger models and towards smaller, more accessible ones with comparable use.

\begin{figure}[h]
    \centering
    \includegraphics[width=\columnwidth]{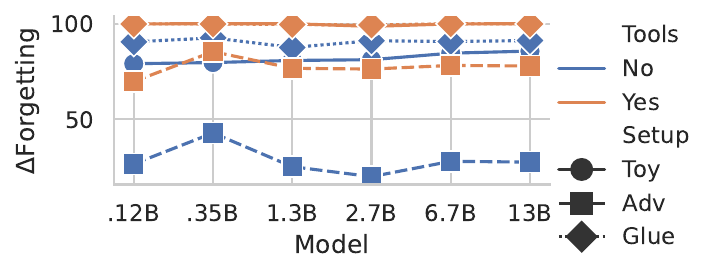}
    \caption{The replay buffer plays a significant role in aiding LMs across tasks in mitigating the forgetting.}
    \label{fig:delta-forgetting}
\end{figure}

\paragraph{Tools can be beneficial for CL.}
By adopting more wide-spread techniques from continual learning, tool LLMs display significant advantages over prototypical LLMs. In particular, by using replay buffer, we observe that forgetting is alleviated to a significantly higher degree when learning with tools. In \autoref{fig:delta-forgetting}, we observe that forgetting drops by $\geq 70\%$ in all tasks.
By comparison, forgetting remains in the $5$-$15$\% range for arithmetic tasks and $\sim 2$\% for the GLUE task when not using tools (as observed in \autoref{fig:Forgetting-log-all-tasks} in \autoref{app:details}), which are all greater than $10\times$ the amount of forgetting that occurs with tools and replay. Though we observe that tool LLMs forget more than vanilla LLMs without replay, the amount of forgetting remains significant (over $80$\%, $30$\% and $85$\% for the three tasks) and limits their practical viability.
What remains important, however, is that models appear capable of learning tools to a much greater capacity, shown by superior learning accuracy throughout. 
These benefits can be observed when using replay (\textcolor{orange}{line} in \autoref{fig:master-results}), where we note the models learn to use the tools almost perfectly,
and the tool LLM can significantly outperform vanilla LLMs in our arithmetic tasks. Even in the case of the more nuanced GLUE task, where the tool is not always correct, benefits are still visible as errors in the final answer result only from the imperfections with the tool, which we can remark due to the fact that the API call accuracy is perfect in these scenarios (see \autoref{app:details}).
These observations bring us to hypothesize that through tool use, LLMs become better at utilizing their parametric knowledge, leading to greater task transfer during CL and allowing them to adapt more effectively.
\section{Discussion}

\paragraph{Parametric Knowledge Utilization.} Studies into language models have shown that pre-training data is oftentimes directly available within trained parameters~\citep{brown2020language, jiang-etal-2020-know, qin-eisner-2021-learning} as \textit{parametric knowledge}. 
However, if the knowledge stored is very example dependent, then it is likely not usable~\cite{kazemnejad2023measuring} in many instances, as there is no direct link between the context in which the knowledge was seen and other examples which are presented to the model~\citep{prato-etal-2023-epik}. As such, one may question whether this knowledge space could be better used.

In contrast, tool learning can generalize the output space, as the learned samples can be more clearly separated into categories based on the tools that are used to solve them. This can make it easier to understand how to handle individual examples from the model perspective and maintain some memory of prior tasks. These observations can explain many of our results, such as improved learning accuracy but greater forgetting when learning tools without replay. If answers are all either numerical values or similar natural language words, there possibly exists a smaller distribution shift that occurs when moving from one task to another. As a result, over-fitting to the answer format may result in a smaller performance degradation. 



\paragraph{Using Auxiliary Expert Systems.} 
Tool LLMs assume that the tools themselves are accurate for the task of interest as otherwise it's existence would be meaningless. But teaching LLMs to make use of tools as auxiliary systems remains a nuanced process; how does it know when to trust the system and take the system response as the truth? There is often a trade-off that exists between speed and performance in these cases; the faster we want the response to be then the more trust we must place in the system to be accurate and not double-guess it. 

Tool LLMs can further be seen as an alternative to mixture of expert models~\citep{jacobs1991localexperts, shazeer2017outrageously, fedus2022review}, which route examples to different experts. However, one can view tool LLMs as a case where the expert exists externally; this leads to a system that may be less coupled with the task.

However, introducing auxiliary systems bring about additional questions. For example, how do we ensure that the model can continuously maintain the ability to use the system properly? 
How is the knowledge for using tools stored and what does it inform us about how much the LLM knows about the tool?
These require further analysis which are necessary both for practical use as well as for understanding LLMs in general.

\paragraph{The Dual Nature of Forgetting.} 
Forgetting is a natural phenomenon, both in humans~\citep{wang2020microglia} and neural networks~\citep{french1999catastrophic}. While it is commonly agreed upon that a symbiotic relationship exists between learning and forgetting within humans~\citep{bjork1970spacing, bjork2019forgetting, gravitz2019importance}, forgetting is still treated as the cause of various failure modes within machine learning~\citep{mccloskey1989catastrophic, ratcliff1990connectionist}. However works have began to show how forgetting and learning can work together symbiotically~\citep{zhou2022fortuitous}.

Forgetting is deemed a negative phenomena which hinders models. However, in the real world, this assessment may not hold in many settings. Recall that updating models with ease is important. For this, unnecessary information should be forgotten as quickly as new information is learnt. This shows that forgetting is not a simple black-or-white issue. When information can become out-dated or incorrect, it may be the case that forgetting is desirable, given that it is no longer useful. Therefore, tool-based models displaying higher forgetting but greater learning accuracy may in fact be desirable, as it demonstrates that models can maintain an ability to learn new information but simultaneously discard information that is no longer relevant. 

\section{Conclusion}

In this work, we explore the potential use of tools in continual learning for LLMs. We apply this setup within a simple arithmetic reasoning setup, where a language model is taught multiple math functions. Our results demonstrate that LLMs that learning to generate answers based on tools both adapt faster to new tasks while also maintaining greater performance on prior tasks. We further validate these conclusions through a continual learning benchmark based on the GLUE natural language understanding benchmark. However, continual learning remains unsolved, as cases still exist where all models fail to demonstrate the ability to autonomously solve the benchmark. This emphasizes the need for models which can adapt to the world in the same manner as conscious humans and by highlighting current limitations and the potential for tool LLMs in this setting, these results hopefully delineate paths for future research which can allow for more practical LLMs deployed in the real world.

\section{Limitations}

Some limitations of this work can be noted in the simplicity of the tools that are explored as well as the degree of relatedness that exists between each tool and how they are used.


First, we note that there exists some relatedness between a number of different functions which we learn due to the granularity at which they are used, which may or may not have resulted in some potential benefits in terms of avoiding catastrophic forgetting. We maintain, however, that we provide enough experimental analysis and results such that this should not pose an issue with the results, hence we believe this to have had minimal effect on potentially producing optimistic results.

Second, forgetting~\citep{chaudhry2018riemannian} is a limited metric, as the concept of `more' forgetting is not well-defined. For example, suppose we take accuracy as our performance metric and are comparing two scenarios. Scenario A has performance degrade by 25\% from a peak performance of 80\% (80\% $\rightarrow$ 60\%). Scenario B observes a 10\% performance degradation from a peak performance of 30\% (30\% $\rightarrow$ 27\%). In this case, despite scenario B observing less forgetting, we may consider it more problematic as the performance was initially significantly worse than A. As such, developing better metrics for capturing these types of phenomena is important for better analysis within continual learning settings. Additionally, as discussed within the paper, it is unclear if zero forgetting is in fact a desirable property and to what extent this metric is able to capture robustness in learning.

\section{Ethical Considerations}

Though not tested for explicitly, there are some negative use-cases that can be derived from this work. Since our results indicate that using tools can perhaps have beneficial properties in terms of avoiding forgetting, this can produce some un-ethical results if the tools themselves are not designed in a way that promotes fairness and proper usage.

For example, suppose one trains a language model to use a generic search tool. If neither the language model nor the tool is designed to avoid searching for potentially harmful or inflammatory material, this runs the risk of potentially being able to propagate false or insensitive information for longer periods of time, as the general ability to make these problematic searches may be retained for longer periods of time.

We also note the computational cost of running the experiments used for this analysis. For transparency, we trained all models on a single machine with 4 NVIDIA V100 32GB GPUs.
\section{Acknowledgements}

Jerry Huang received partial financial support from the National Science and Engineering Research Council (NSERC), the Fonds de Recherche du Qu\'{e}bec Nature et technologies (FRQNT), Hydro-Qu\'{e}bec as well as a graduate bursary from the Universit\'{e} de Montr\'{e}al. Sarath Chandar is supported by a Canada CIFAR AI Chair, a Canada Research Chair in Lifelong Machine Learning and a NSERC Discovery Grant. The authors thank the NLP team at Huawei Noah's Ark Lab for productive discussion and technical support. The authors also thank Andreas Madsen, Pranshu Malviya, Janarthanan Rajendran and Gon\c{c}alo Mordido for useful comments during the initial conception of this work.

\bibliography{anthology, custom, other}

\appendix

\section{Implementation Details}
\label{app:hardware}

\begin{table}[ht]
    \centering
    \resizebox{\linewidth}{!}{%
    \begin{tabular}{cccc}
        \toprule
         \textsc{Model} & \textsc{Toy} & \textsc{Advanced} & \textsc{GLUE} \\
         \midrule
         \texttt{OPT-125M}& $15$ min & $30$ min & $25$ min \\
         \texttt{OPT-350M}& $15$ min & $30$ min & $25$ min \\
         \texttt{OPT-1.3B}& $2$ h & $4$ h & $3$ h $20$ min \\
         \texttt{OPT-2.7B}& $3$ h & $6$ h $30$ min & $5$ h \\
         \texttt{OPT-6.7B}& $4$ h & $8$ min & $7$ h $20$ min \\
         \texttt{OPT-13B}& $12$ h & $24$ h & $21$ h $30$ min \\
         \bottomrule
    \end{tabular}
    }
    \caption{Approximate training times for each model.}
    \label{tab:times}
\end{table}

For the experiments in this study, we exclusively
use a server of 4 NVIDIA V100-32GB GPUs. Models with $\leq 350$M parameters were trained on a single GPU. Other models used the entire server of 4 GPUs. The approximate training time (rounded to the nearest hour) for a single seed for each model without replay are presented in \autoref{tab:times}. Times with replay require approximately twice the amount of computation time.

\section{Benchmarks}
\label{app:data-info}

\subsection{Dataset Size}

\paragraph{Toy Arithmetic Benchmark.} We generate an example for each function and operand. This results in 10000 examples per task, which we separate into a train and test set using an 80\%-20\% random split.

\paragraph{Advanced Arithmetic Benchmark.} We generate 20000 examples per task, which we again separate into a train and test set using an 80\%-20\% random split.

\paragraph{GLUE Benchmark.} Due to different tasks having different sizes, we use the size of the smallest task as our limits for both the train and test set for each task. In our case, we use 8.5k example from the training set of each task and 5k examples from the test set of each task. 

\subsection{Templates}
Templates for our benchmarks are in \autoref{tab:math-templates} and \autoref{tab:glue-templates}.

\begin{table}[th]
    \centering
    \resizebox{\linewidth}{!}{%
    \begin{tabular}{ll}
        \toprule
        \textbf{Description} & \textbf{Templates} \\
        \midrule
        \multirow{4}{*}{Addition} & What is $a$ plus $b$?\\
        & What is the sum of $a$ and $b$?\\
        & What do you get if you add $a$ to $b$?\\
        & What do the total of $a$ and $b$?\\
        \midrule
        \multirow{5}{*}{Subtraction} & What is $a$ minus $b$?\\
        & What is the difference between $a$ and $b$? \\
        & What is $b$ less than $a$? \\
        & What is $a$ take away $b$? \\
        & What is the distance between $a$ and $b$? \\
        \midrule
        \multirow{4}{*}{Multiplication} & What is $a$ times $b$?\\
        & What is the product of $a$ and $b$?\\
        & How much is $a$ groups of $b$?\\
        & $a$ multiples of $b$ is how much?\\
        \midrule
        \multirow{4}{*}{Division} & What is $a$ divided by $b$?\\
        & What is the quotient of $a$ and $b$?\\
        & How many times does $b$ fit into $a$?\\
        & How $b$ go into $a$?\\
        \midrule
        \multirow{3}{*}{GCD} & What is the greatest common factor of $a$ and $b$?\\
        & Calculate the highest common divisor of $a$ and $b$.\\
        & What is the largest number that divides both $a$ and $b$.\\
        \midrule
        \multirow{2}{*}{LCM} & What is the smallest common multiple of $a$ and $b$?\\
        & What is the smallest number that is a multiple of both $a$ and $b$.\\
        \midrule
        \multirow{2}{*}{Prime Factors} & What are the prime factors of $a$?\\
        & Which factors of $a$ are prime?\\
        \bottomrule
    \end{tabular}}
    \caption{Templates for arithmetic tools.}
    \label{tab:math-templates}
\end{table}

\begin{table*}[th]
    \centering
    \resizebox{\linewidth}{!}{%
    \begin{tabular}{llr}
        \toprule
        \textbf{Task} & \textbf{Templates} & \textbf{Answers} \\
        \midrule
        \multirow{2}{*}{MNLI} & Does \texttt{[Sentence 1]} either entail or contradict \texttt{[Sentence 2]}, or neither? & \multirow{2}{*}{\{entailment, contradiction, neutral\}}\\
        & Is \texttt{[Sentence 2]} an entailment or contradiction of \texttt{[Sentence 1]}, or neither? & \\
        \midrule
        \multirow{2}{*}{QQP} & Is \texttt{[Sentence 1]} a paraphrasing of \texttt{[Sentence 2]}? & \multirow{2}{*}{\{yes, no\}}\\
        & Are \texttt{[Sentence 1]} and \texttt{[Sentence 2]} paraphrases? &\\
        \midrule
        \multirow{2}{*}{CoLA} & Is \texttt{[Sentence 1]} a linguistically acceptable sentence? &  \multirow{2}{*}{\{yes, no\}}\\
        & Does \texttt{[Sentence 1]} make sense linguistically? & \\
        \midrule
        \multirow{2}{*}{SST-2} & Does \texttt{[Sentence 1]} express a positive or negative sentiment? & \multirow{2}{*}{\{positive, negative\}}\\
        & What kind of sentiment does \texttt{[Sentence 1]} show? & \\
        \bottomrule
    \end{tabular}}
    \caption{Templates for GLUE tools.}
    \label{tab:glue-templates}
\end{table*}

\section{Training Hyperparameters}
\label{app:hyperparameters}
We use the following hyperparameters for training a model on each task within the continual learning setup.

\begin{table}[ht]
    \centering
    \resizebox{\linewidth}{!}{
    \begin{tabular}{lr}
        \toprule
        \textbf{Hyperparameter} & \textbf{Value} \\
        \midrule
        Batch Size & $64$ \\
        \multirow{2}{*}{Peak Learning Rate} & If (\# of parameters $< 10^9$): $2\times 10^{-5}$ \\
        & else: $1\times 10^{-5}$ \\
        Weight Decay & $0.01$ \\
        Adam $\epsilon$ & $1\times 10^{-8}$ \\
        Adam $\beta_1$ & $0.9$ \\
        Adam $\beta_2$ & $0.99$ \\
        Learning Rate Scheduler & Linear Warm-Up and Decay \\
        Warm-Up (Per Task) & $10\%$ of training steps \\
        Gradient Clipping & $0.0$ \\
        \bottomrule
    \end{tabular}
    }
    \caption{Hyper-parameters for learning each tool.}
    \label{tab:hyperparameters}
\end{table}

\subsection{Training Task Sequences}

In continual learning, seeding occurs across the task sequence rather than the individual training examples per sequence. For example, given a set of tasks $\mathcal{T} = \{\mathcal{T}_1, \mathcal{T}_2, \dots, \mathcal{T}_T\}$, the training can occur over any permutation of the task ordering.

\section{Evaluation Metrics}\label{app:eval}


\paragraph{Average Performance} We denote the performance on the test set of task $\tau$ when the model is being trained on task $T$ as $p_{T, \tau}$. The average performance across all learned tasks is therefore
\[
\bar{p} = \frac{1}{T} \sum_{\tau=1}^{T} p_{T, \tau}
\]

\paragraph{Forgetting~\citep{chaudhry2018riemannian}} Supposing that the model is being trained on task $T$. Forgetting is defined as the amount by which performance on all previous tasks has degraded from it's previous maximum performance. If the performance metric is defined such that increased performance corresponds to an increase in metric value, then forgetting can be calculated as
\[
F_{T} = \frac{1}{T-1}  \sum_{\tau=1}^{T-1} \underset{t \in \{1,\dots T-1\}}{\text{max}}(p_{t, \tau} - p_{T, \tau})
\]
Otherwise, if the performance metric is defined such that increased performance corresponds to an increase in metric value, then forgetting can be calculated as
\[
F_{T} = \frac{1}{T-1}  \sum_{\tau=1}^{T-1} \underset{t \in \{1,\dots T-1\}}{\text{min}}(p_{T, \tau} - p_{t, \tau})
\]
However, this can converted to a relative forgetting over tasks
\[
F_{T} = \frac{1}{T-1}  \sum_{\tau=1}^{T-1} \underset{t \in \{1,\dots T-1\}}{\text{max}}\bigg(\frac{p_{t, \tau} - p_{T, \tau}}{p_{t, \tau}}\bigg)
\]
to account for general performance differences between tasks.

To further illustrate this, again suppose a sequence of task being presented to the model, $\{\mathcal{T}_1, \dots, \mathcal{T}_n\}$. The model has been trained on $\mathcal{T}_1$ and $\mathcal{T}_2$. If the model originally achieved an accuracy of 85\% on $\mathcal{T}_1$ after training on $\mathcal{T}_1$ and this number drops to 50\% after training on $\mathcal{T}_2$, we would say that there is a degradation of performance of 35\%. However, in order to account for the fact that peak performance on tasks can differ, we normalize this value using the highest achieved accuracy on each task. As such, forgetting is presented as a normalized value between 0 and 1, representing the average degradation from peak performance across all tasks. This is only calculated from the second task onwards.

\paragraph{Learning Capacity~\citep{riemer2018learning}} Learning capacity is defined as the ability to learn a new task immediately as the model sees it. Therefore it can be defined as
\[
L_{T} = {\frac{1}{T} \sum_{\tau=1}^T p_{\tau,\tau}}
\]
where $p_{i,j}$ measures the accuracy on the test set of task $j$ while learning task $i$.

As an example, again suppose a sequence of tasks $\{\mathcal{T}_1, \dots, \mathcal{T}_n\}$ are presented to a model. After training on $\mathcal{T}_1$, we record the average performance on $\mathcal{T}_1$'s test set and report this as the learning accuracy. After training on $\mathcal{T}_2$, we record performance on the test set for $\mathcal{T}_2$, which we then average with the previously calculated immediate accuracy after $\mathcal{T}_1$ to record the learning accuracy after $\mathcal{T}_2$.

\section{Tools}
\label{app:tools}

\subsection{Evaluating API Calls}
In order to evaluate API calls, we create a tool which can parse the generated calls. For example, if the generated call is `\textsc{ADD(5, 5)}', then use a regular expression parser to separate `\textsc{ADD}' as the function and `\textsc{5, 5}' as the arguments passed to the function. We then take this parsed answer, feed it to a pre-written function that can evaluate it and then compare it to the raw numerical answer.

\subsection{Building our tools}

For our arithmetic tasks, we explicitly design functions which take integer arguments. As an example, we simply build our function for addition as
\begin{verbatim}
def add(a, b):
    return a + b
\end{verbatim}
with the output being directly compared with the label provided in the dataset.

For the continual GLUE tasks, we instead use a frozen LLM available from HuggingFace that is meant to produce an answer, for example
\begin{verbatim}
from transformers import pipeline
def entailment(sentence, model=None):
    pipe = pipeline(model=model)
    return pipe(sentence)[0]['label']
\end{verbatim}
after which we again compare the returned label with the ground truth provided in the dataset. By default, we use publicly available base-sized RoBERTa models that have already been fine-tuned on the corresponding task as the tool model.

\section{Detailed results}\label{app:details}

The following sections provide some additional numerical details and results from our training.

\subsection{Mixed Training}

Comprehensive results for models on our benchmarks during mixed training are shown in \autoref{tab:mixed-arithmetic} and \autoref{tab:mixed-glue}. Results are all averaged over 10 seeds. We omit standard error values as they are all $\sigma \leq 0.1$ in each case.

\begin{table}[ht!]
    \centering
    \resizebox{\linewidth}{!}{
        \begin{tabular}{l|ccc|ccc}
            \toprule
            \multicolumn{1}{c}{\textbf{Model}} & \multicolumn{3}{c}{\textbf{Toy}} & \multicolumn{3}{c}{\textbf{Advanced}} \\
            \multicolumn{1}{c}{\textit{Tools}$\rightarrow$} & \textbf{No} & \textbf{Yes} & \multicolumn{1}{c}{($\uparrow\downarrow$)} & \textbf{No} & \textbf{Yes} & \multicolumn{1}{c}{($\uparrow\downarrow$)}\\
            \midrule
            \texttt{OPT-125M} 
                & 44.4 & \phantom{0}99.9 & \ua{55.5} 
                & 16.9 & 97.1 & \ua{80.2} \\
            \texttt{OPT-350M} 
                & 43.3 & 100.0 & \ua{56.7} 
                & 17.8 & 97.6 & \ua{79.8} \\
            \texttt{OPT-1.3B} 
                & 91.3 & 100.0 & \ua{8.6}\phantom{0}
                & 36.7 & 97.9 & \ua{61.2} \\
            \texttt{OPT-2.7B} 
                & 95.5 & 100.0 & \ua{4.5}\phantom{0}
                & 41.9 & 99.3 & \ua{57.4} \\
            \texttt{OPT-6.7B} 
                & 98.0 & 100.0 & \ua{2.0}\phantom{0} 
                & 46.2 & 99.1 & \ua{52.9} \\
            \texttt{OPT-13B} 
                & 99.0 & 100.0 & \ua{1.0}\phantom{0} 
                & 50.0 & 99.4 & \ua{49.4} \\
            \bottomrule
        \end{tabular}
    }
    \caption{Accuracy of models on both benchmarks when data is mixed into a single task. These represent upper bounds for performance. 
    Colored boxes refer to the increase or decrease in accuracy performance from the non-tool to tool setup, with green representing a positive change and red a negative change.
    }
    \label{tab:mixed-arithmetic}
\end{table}

\begin{table}[ht!]
    \centering
    \resizebox{\linewidth}{!}{%
        \begin{tabular}{lccc}
            \toprule
            \multicolumn{1}{c}{\textbf{Model}} & \textbf{Raw Ans.} & \textbf{API Call} & \textbf{API Ans.} \\
            \midrule
            \texttt{OPT-125M} 
            & 69.7 & \phantom{0}99.9 & 91.4 \ua{21.57} \\
            \texttt{OPT-350M} 
            & 74.6 & \phantom{0}99.9 & 91.4 \ua{16.8}\phantom{0} \\
            \texttt{OPT-1.3B} 
            & 80.4 & 100.0 & 91.4 \ua{11.0}\phantom{0} \\
            \texttt{OPT-2.7B}
            & 82.0 & 100.0 & 91.4 \ua{9.4}\phantom{00} \\
            \texttt{OPT-6.7B} 
            & 87.5 & 100.0 & 91.4 \ua{3.9}\phantom{00} \\
            \texttt{OPT-13B} 
            & 88.2 & 100.0 & 91.4 \ua{3.2}\phantom{00} \\
            \bottomrule
        \end{tabular}
    }
    \caption{Performance of models on our continual GLUE setup when tasks are mixed. Comparison between non-tool LLMs and tool LLMs is done using the tool output.}
    \label{tab:mixed-glue}
\end{table}

\subsection{Sequential Training}
Comprehensive results for models on our benchmarks during sequential training are shown in \autoref{tab:cl-arithmetic-se} and \autoref{tab:cl-glue-se}. Results are all averaged over 10 seeds. We report standard errors for all performance metrics.

\begin{table*}[ht!]
\centering
    \resizebox{\linewidth}{!}{%
    \begin{tabular}{clcccccc}
        \toprule
        & \textbf{\textsc{Model}} 
        & \multicolumn{3}{c}{\textbf{\textsc{Without Tools}}} 
        & \multicolumn{3}{c}{\textbf{\textsc{With Tools}}} \\
        & 
        & \textsc{Acc.} & \textsc{For.} & \textsc{L-A.}
        & \textsc{Acc.} & \textsc{For.} & \textsc{L-A.}\\
        \midrule
        \parbox[t]{2mm}{\multirow{12}{*}{\rotatebox[origin=c]{90}{\textbf{Toy Benchmark}}}} 
        & \texttt{OPT-125M} 
            & $\phantom{0}8.6$ 
            & $84.9$ 
            & $24.7$ 
            & $\phantom{0}25.0$ \ua{16.4} 
            & $100.0$ \ub{15.1}
            & $\phantom{0}99.8$ \ua{75.1} \\
        & \texttt{OPT-350M} 
            & $\phantom{0}9.0$ 
            & $92.9$ 
            & $25.7$ 
            & $\phantom{0}25.0$ \ua{16.0}
            & $100.0$ \ub{7.1}\phantom{0} 
            & $\phantom{0}99.8$ \ua{74.1} \\
        & \texttt{OPT-1.3B} 
            & $30.3$ 
            & $89.7$ 
            & $83.1$ 
            & $\phantom{0}25.0$ \da{5.3}\phantom{0} 
            & $100.0$ \ub{10.3} 
            & $100.0$ \ua{16.9} \\
        & \texttt{OPT-2.7B} 
            & $31.1$ 
            & $89.9$ 
            & $89.4$
            & $\phantom{0}26.0$ \da{5.1}\phantom{0}
            & $\phantom{0}98.6$ \ub{8.8}\phantom{0} 
            & $100.0$ \ua{10.6} \\
        & \texttt{OPT-6.7B}
            & $26.0$ 
            & $97.1$
            & $93.0$ 
            & $25.0$ \da{1.0}
            & $100.0$ \ub{2.9}\phantom{0}
            & $100.0$ \ua{7.0}\phantom{0} \\
        & \texttt{OPT-13B} 
            & $28.4$ 
            & $98.1$
            & $95.3$ 
            & $25.0$ \da{1.0} 
            & $100.0$ \ub{1.9}\phantom{0} 
            & $100.0$ \ua{4.7}\phantom{0} \\
        \cmidrule{2-8}
        & \texttt{OPT-125M}\textsubscript{\texttt{+ER}} 
            & $36.9$ 
            & $\phantom{0}5.8$ 
            & $26.0$ 
            & $\phantom{0}99.9$ \ua{63.0} 
            & $\phantom{0}\phantom{0}0.1$ \db{5.7}\phantom{0} 
            & $\phantom{0}99.8$ \ua{73.8} \\
        & \texttt{OPT-350M}\textsubscript{\texttt{+ER}} 
            & $33.7$ 
            & $13.2$ 
            & $27.5$
            & $\phantom{0}99.9$ \ua{66.2} 
            & $\phantom{0}\phantom{0}0.0$ \db{13.2} 
            & $\phantom{0}99.8$ \ua{72.2} \\
        & \texttt{OPT-1.3B}\textsubscript{\texttt{+ER}} 
            & $80.8$ 
            & $\phantom{0}8.9$ 
            & $81.8$ 
            & $100.0$ \ua{9.2}\phantom{0}
            & $\phantom{0}\phantom{0}0.0$ \db{2.9}\phantom{0}
            & $100.0$ \ua{18.2} \\
        & \texttt{OPT-2.7B}\textsubscript{\texttt{+ER}}
            & $83.6$ 
            & $\phantom{0}8.7$ 
            & $89.5$ 
            & $100.0$ \ua{6.4}\phantom{0} 
            & $\phantom{0}\phantom{0}0.0$ \db{1.7}\phantom{0} 
            & $100.0$ \ua{10.5} \\
        & \texttt{OPT-6.7B}\textsubscript{\texttt{+ER}}
            & $87.9$
            & $12.5$
            & $93.8$
            & $100.0$ \ua{12.1} 
            & $\phantom{0}\phantom{0}0.1$ \db{12.4} 
            & $100.0$ \ua{6.2}\phantom{0} \\
        & \texttt{OPT-13B}\textsubscript{\texttt{+ER}} 
            & $89.8$
            & $12.4$ 
            & $94.9$ 
            & $100.0$ \ua{10.2} 
            & $\phantom{0}\phantom{0}0.0$ \db{12.4}
            & $100.0$ \ua{5.1}\phantom{0} \\
        \midrule
        \parbox[t]{2mm}{\multirow{12}{*}{\rotatebox[origin=c]{90}{\textbf{Advanced Benchmark}}}} 
        & \texttt{OPT-125M}   
            & $10.3$ 
            & $39.3$ 
            & $15.8$ 
            & $\phantom{0}38.4$ \ua{28.1} 
            & $\phantom{0}70.8$ \ub{31.5} 
            & $\phantom{0}95.1$ \ua{79.3} \\
        & \texttt{OPT-350M}
            & $\phantom{0}7.0$ 
            & $55.9$ 
            & $15.5$ 
            & $\phantom{0}25.6$ \ua{18.7} 
            & $\phantom{0}85.8$ \ub{29.9} 
            & $\phantom{0}96.6$ \ua{81.1} \\
        & \texttt{OPT-1.3B}
            & $25.4$ 
            & $36.9$ 
            & $35.6$ 
            & $\phantom{0}33.4$ \ua{8.0}\phantom{0} 
            & $\phantom{0}77.2$ \ub{40.3} 
            & $\phantom{0}98.5$ \ua{62.9} \\
        & \texttt{OPT-2.7B}
            & $32.1$ 
            & $33.2$ 
            & $41.4$ 
            & $\phantom{0}33.7$ \ua{1.6}\phantom{0} 
            & $\phantom{0}76.9$ \ub{48.6} 
            & $\phantom{0}98.7$ \ua{57.3} \\
        & \texttt{OPT-6.7B}
            & $31.6$ 
            & $40.6$ 
            & $47.3$ 
            & $\phantom{0}32.2$ \ua{0.7}\phantom{0} 
            & $\phantom{0}78.7$ \ub{38.0} 
            & $\phantom{0}98.9$ \ua{51.7} \\
        & \texttt{OPT-13B}
            & $32.5$ 
            & $39.5$ 
            & $47.2$ 
            & $\phantom{0}33.2$ \ua{0.7}\phantom{0} 
            & $\phantom{0}78.4$ \ub{38.0} 
            & $\phantom{0}99.0$ \ua{51.8} \\
        \cmidrule{2-8}
        & \texttt{OPT-125M}\textsubscript{\texttt{+ER}}
            & $15.5$ 
            & $12.6$ 
            & $16.2$ 
            & $\phantom{0}96.1$ \ua{80.6} 
            & $\phantom{00}1.0$ \db{11.6}
            & $\phantom{0}92.8$ \ua{76.6} \\
        & \texttt{OPT-350M}\textsubscript{\texttt{+ER}}    
            & $15.4$ 
            & $13.0$ 
            & $16.3$ 
            & $\phantom{0}96.8$ \ua{81.4} 
            & $\phantom{00}0.4$ \db{12.6} 
            & $\phantom{0}95.6$ \ua{79.3} \\
        & \texttt{OPT-1.3B}\textsubscript{\texttt{+ER}}    
            & $34.5$ 
            & $11.6$ 
            & $35.1$ 
            & $\phantom{0}96.6$ \ua{62.1} 
            & $\phantom{00}0.5$ \db{11.1} 
            & $\phantom{0}97.9$ \ua{61.8}\\
        & \texttt{OPT-2.7B}\textsubscript{\texttt{+ER}}    
            & $38.9$ 
            & $13.1$ 
            & $39.7$ 
            & $\phantom{0}97.4$ \ua{58.5} 
            & $\phantom{00}0.6$ \db{12.5} 
            & $\phantom{0}97.7$ \ua{57.0}\\
        & \texttt{OPT-6.7B}\textsubscript{\texttt{+ER}}    
            & $41.6$ 
            & $12.5$ 
            & $42.1$ 
            & $\phantom{0}97.8$ \ua{56.6} 
            & $\phantom{00}0.5$ \db{12.0} 
            & $\phantom{0}98.7$ \ua{55.6}\\
        & \texttt{OPT-13B}\textsubscript{\texttt{+ER}}    
            & $42.4$ 
            & $11.8$ 
            & $44.2$ 
            & $\phantom{0}97.7$ \ua{56.3} 
            & $\phantom{00}0.5$ \db{11.3}  
            & $\phantom{0}98.9$ \ua{54.7}\\
        \bottomrule
    \end{tabular}
    }
    \caption{Performance on the continual learning version of our arithmetic benchmarks. Performance is noted in terms of accuracy (Acc.), forgetting (For.) and learning accuracy (L-A.), both when learning with or without tools (averaged across 10 seeds). $\uparrow$ and $\downarrow$ indicate direction of performance change when using tools, with green/red indicating if the change is desirable or undesirable. {\texttt{+ER}} indicates models use a replay buffer of 64 samples times the number of tasks. 
    }
    \label{tab:cl-arithmetic}
\end{table*}

\begin{table*}[ht!]
\centering
\resizebox{\linewidth}{!}{
    \begin{tabular}{clcccccc}
        \toprule
        & \textbf{\textsc{Model}} 
        & \multicolumn{3}{c}{\textbf{\textsc{Without Tools}}} 
        & \multicolumn{3}{c}{\textbf{\textsc{With Tools}}} \\
        & 
        & {\textsc{Acc. ($\uparrow$)}} & {\textsc{For. ($\downarrow$)}} & \textsc{L-A. ($\uparrow$)}
        & {\textsc{Acc. ($\uparrow$)}} & {\textsc{For. ($\downarrow$)}} & \textsc{L-A. ($\uparrow$)}\\
        \midrule
        \parbox[t]{2mm}{\multirow{12}{*}{\rotatebox[origin=c]{90}{\textbf{Toy Benchmark}}}} 
        & \texttt{OPT-125M} 
            & $\phantom{0}8.6_{\pm 1.0}$ 
            & $84.9_{\pm 1.7}$ 
            & $24.7_{\pm 0.1}$ 
            & $25.0_{\pm 0.0}$
            & $100.0_{\pm 0.0}$
            & $\phantom{0}99.8_{\pm 0.0}$ \\
        & \texttt{OPT-350M} 
            & $\phantom{0}9.0_{\pm 1.0}$ 
            & $92.9_{\pm 0.9}$ 
            & $25.7_{\pm 0.1}$ 
            & $25.0_{\pm 0.0}$
            & $100.0_{\pm 0.0}$
            & $\phantom{0}99.8_{\pm 0.0}$ \\
        & \texttt{OPT-1.3B} 
            & $30.3_{\pm 0.6}$ 
            & $89.7_{\pm 0.8}$ 
            & $83.1_{\pm 0.8}$ 
            & $25.0_{\pm 0.0}$
            & $100.0_{\pm 0.0}$ 
            & $100.0_{\pm 0.0}$ \\
        & \texttt{OPT-2.7B} 
            & $31.1_{\pm 0.7}$ 
            & $89.9_{\pm 1.1}$ 
            & $89.4_{\pm 0.5}$
            & $26.0_{\pm 0.3}$
            & $\phantom{0}98.6_{\pm 0.4}$
            & $100.0_{\pm 0.0}$ \\
        & \texttt{OPT-6.7B}
            & $26.0_{\pm 0.4}$ 
            & $97.1_{\pm 0.5}$
            & $93.0_{\pm 0.2}$ 
            & $25.0_{\pm 0.0}$
            & $100.0_{\pm 0.0}$
            & $100.0_{\pm 0.0}$ \\
        & \texttt{OPT-13B} 
            & $28.4_{\pm 0.3}$ 
            & $98.1_{\pm 0.4}$
            & $95.3_{\pm 0.1}$ 
            & $25.0_{\pm 3.4}$
            & $100.0_{\pm 0.0}$
            & $100.0_{\pm 0.0}$ \\
        \cmidrule{2-8}
        & \texttt{OPT-125M}\textsubscript{\texttt{+ER}} 
            & $36.9_{\pm 0.6}$ 
            & $\phantom{0}5.8_{\pm 1.7}$ 
            & $26.0_{\pm 0.4}$ 
            & $\phantom{0}99.9_{\pm 0.0}$
            & $\phantom{0}\phantom{0}0.1_{\pm 0.0}$ 
            & $\phantom{0}99.8_{\pm 0.0}$ \\
        & \texttt{OPT-350M}\textsubscript{\texttt{+ER}} 
            & $33.7_{\pm 0.5}$ 
            & $13.2_{\pm 1.8}$ 
            & $27.5_{\pm 0.2}$
            & $\phantom{0}99.9_{\pm 0.0}$
            & $\phantom{0}\phantom{0}0.0_{\pm 0.0}$
            & $\phantom{0}99.8_{\pm 0.1}$ \\
        & \texttt{OPT-1.3B}\textsubscript{\texttt{+ER}} 
            & $80.8_{\pm 0.5}$ 
            & $\phantom{0}8.9_{\pm 0.5}$ 
            & $81.8_{\pm 2.5}$ 
            & $100.0_{\pm 0.0}$
            & $\phantom{0}\phantom{0}0.0_{\pm 0.0}$
            & $100.0_{\pm 0.0}$ \\
        & \texttt{OPT-2.7B}\textsubscript{\texttt{+ER}}
            & $83.6_{\pm 0.4}$ 
            & $\phantom{0}8.7_{\pm 0.3}$ 
            & $89.5_{\pm 1.1}$ 
            & $100.0_{\pm 0.0}$
            & $\phantom{0}\phantom{0}0.0_{\pm 0.0}$
            & $100.0_{\pm 0.0}$ \\
        & \texttt{OPT-6.7B}\textsubscript{\texttt{+ER}}
            & $87.9_{\pm 0.8}$
            & $12.5_{\pm 0.9}$
            & $93.8_{\pm 0.5}$
            & $100.0_{\pm 0.0}$
            & $\phantom{0}\phantom{0}0.1_{\pm 0.0}$
            & $100.0_{\pm 0.0}$ \\
        & \texttt{OPT-13B}\textsubscript{\texttt{+ER}} 
            & $89.8_{\pm 0.7}$
            & $12.4_{\pm 1.1}$ 
            & $94.9_{\pm 0.6}$ 
            & $100.0_{\pm 0.0}$
            & $\phantom{0}\phantom{0}0.0_{\pm 0.0}$
            & $100.0_{\pm 0.0}$ \\
        \midrule
        \parbox[t]{2mm}{\multirow{12}{*}{\rotatebox[origin=c]{90}{\textbf{Advanced Benchmark}}}} 
        & \texttt{OPT-125M}   
            & $10.3_{\pm 0.7}$ 
            & $39.3_{\pm 7.2}$ 
            & $15.8_{\pm 0.1}$ 
            & $\phantom{0}38.4_{\pm 3.4}$
            & $\phantom{0}70.8_{\pm 4.0}$ 
            & $\phantom{0}95.1_{\pm 0.7}$ \\
        & \texttt{OPT-350M}
            & $\phantom{0}7.0_{\pm 0.7}$ 
            & $55.9_{\pm 6.2}$ 
            & $15.5_{\pm 0.3}$ 
            & $\phantom{0}25.6_{\pm 2.4}$ 
            & $\phantom{0}85.8_{\pm 2.9}$
            & $\phantom{0}96.6_{\pm 0.2}$ \\
        & \texttt{OPT-1.3B}
            & $25.4_{\pm 1.7}$ 
            & $36.9_{\pm 4.8}$ 
            & $35.6_{\pm 0.3}$ 
            & $\phantom{0}33.4_{\pm 3.9}$ 
            & $\phantom{0}77.2_{\pm 4.6}$ 
            & $\phantom{0}98.5_{\pm 0.0}$ \\
        & \texttt{OPT-2.7B}
            & $32.1_{\pm 1.9}$ 
            & $33.2_{\pm 4.2}$ 
            & $41.4_{\pm 0.4}$ 
            & $\phantom{0}33.7_{\pm 4.2}$
            & $\phantom{0}76.9_{\pm 4.9}$
            & $\phantom{0}98.7_{\pm 0.0}$ \\
        & \texttt{OPT-6.7B}
            & $31.6_{\pm 1.8}$ 
            & $40.6_{\pm 4.2}$ 
            & $47.3_{\pm 0.2}$ 
            & $\phantom{0}32.2_{\pm 3.0}$
            & $\phantom{0}78.7_{\pm 3.5}$
            & $\phantom{0}98.9_{\pm 0.0}$ \\
        & \texttt{OPT-13B}
            & $32.5_{\pm 1.2}$ 
            & $39.5_{\pm 3.6}$ 
            & $47.2_{\pm 0.2}$ 
            & $\phantom{0}33.2_{\pm 2.8}$
            & $\phantom{0}78.4_{\pm 3.2}$
            & $\phantom{0}99.0_{\pm 0.0}$ \\
        \cmidrule{2-8}
        & \texttt{OPT-125M}\textsubscript{\texttt{+ER}}
            & $15.5_{\pm 0.3}$ 
            & $12.6_{\pm 2.3}$ 
            & $16.2_{\pm 0.1}$ 
            & $\phantom{0}96.1_{\pm 0.2}$ 
            & $\phantom{00}1.0_{\pm 0.1}$ 
            & $\phantom{0}92.8_{\pm 0.8}$ \\
        & \texttt{OPT-350M}\textsubscript{\texttt{+ER}}    
            & $15.4_{\pm 0.1}$ 
            & $13.0_{\pm 1.4}$ 
            & $16.3_{\pm 0.2}$ 
            & $\phantom{0}96.8_{\pm 0.1}$
            & $\phantom{00}0.4_{\pm 0.1}$ 
            & $\phantom{0}95.6_{\pm 0.3}$ \\
        & \texttt{OPT-1.3B}\textsubscript{\texttt{+ER}}    
            & $34.5_{\pm 0.2}$ 
            & $11.6_{\pm 1.9}$ 
            & $35.1_{\pm 0.2}$ 
            & $\phantom{0}96.6_{\pm 0.2}$ 
            & $\phantom{00}0.5_{\pm 0.1}$
            & $\phantom{0}97.9_{\pm 0.9}$ \\
        & \texttt{OPT-2.7B}\textsubscript{\texttt{+ER}}    
            & $38.9_{\pm 0.1}$ 
            & $13.1_{\pm 1.1}$ 
            & $39.7_{\pm 0.1}$ 
            & $\phantom{0}97.4_{\pm 0.3}$
            & $\phantom{00}0.6_{\pm 0.1}$ 
            & $\phantom{0}97.7_{\pm 0.5}$ \\
        & \texttt{OPT-6.7B}\textsubscript{\texttt{+ER}}    
            & $41.6_{\pm 0.1}$ 
            & $12.5_{\pm 1.4}$ 
            & $42.1_{\pm 0.1}$ 
            & $\phantom{0}97.8_{\pm 0.2}$ 
            & $\phantom{00}0.5_{\pm 0.1}$ 
            & $\phantom{0}98.7_{\pm 0.8}$ \\
        & \texttt{OPT-13B}\textsubscript{\texttt{+ER}}    
            & $42.4_{\pm 0.1}$ 
            & $11.8_{\pm 1.2}$ 
            & $44.2_{\pm 0.1}$ 
            & $\phantom{0}97.7_{\pm 0.2}$
            & $\phantom{00}0.5_{\pm 0.1}$
            & $\phantom{0}98.9_{\pm 0.7}$ \\
        \bottomrule
        \end{tabular}
    }
    \caption{Performance of models on the continual learning version of our arithemtic benchmarks. Given that each tool in these settings are oracles, API call correctness and final answer correctness with tools is the same. Performance is noted in terms of accuracy (Acc.), forgetting (For.) and learning accuracy (L-A.), both when learning with or without tools (averaged across 10 seeds). $\uparrow$ indicates that higher values are better, $\downarrow$ lower. {\texttt{+ER}} indicates models use a replay buffer of 64 samples times the number of tasks. 
    }
    \label{tab:cl-arithmetic-se}
\end{table*}

\begin{table*}[ht]
    \centering
    \resizebox{\linewidth}{!}{%
        \begin{tabular}{lcccccc}
        \toprule
        \multicolumn{1}{c}{\textbf{\textsc{Model}}} & 
        \multicolumn{3}{c}{\textbf{\textsc{Without Tools}}} & 
        \multicolumn{3}{c}{\textbf{\textsc{With Tools}}} \\ 
        & \textsc{Acc.} & \textsc{For.} & \textsc{L-A.} 
        & \textsc{Acc.} & \textsc{For.} & \textsc{L-A.} \\
        \midrule
        \texttt{OPT-125M} 
        & $27.0$ & $92.4$ & $56.5$
        & $23.4$ \da{6.6} & $100.0$ \ub{8.6} & $91.4$ \ua{34.9} \\
        \texttt{OPT-350B} 
        & $27.2$ & $94.4$ & $57.3$
        & $23.3$ \da{6.9} & $100.0$ \ub{5.6} & $91.5$ \ua{34.2} \\
        \texttt{OPT-1.3B} 
        & $27.2$ & $89.7$ & $67.3$
        & $23.4$ \da{3.8} & $\phantom{0}99.6$ \ub{9.9} & $91.4$ \ua{24.1} \\
        \texttt{OPT-2.7B} 
        & $28.9$ & $93.2$ & $69.2$
        & $23.5$ \da{5.4} & $\phantom{0}99.4$ \ub{6.2} & $91.2$ \ua{22.0} \\
        \texttt{OPT-6.7B} 
        & $30.2$ & $92.7$ & $71.7$
        & $23.6$ \da{6.6} & $100.0$ \ub{7.3} & $91.4$ \ua{19.7} \\
        \texttt{OPT-13B} 
        & $29.9$ & $93.4$ & $72.1$
        & $23.5$ \da{6.4} & $100.0$ \ub{6.6} & $91.4$ \ua{19.3} \\
        \midrule
        \texttt{OPT-125M}\textsubscript{\texttt{+ER}} 
        & $60.7$ & $\phantom{0}1.9$ & $62.2$
        & $91.3$ \ua{30.6} & $\phantom{00}0.1$ \db{1.8} & $91.4$ \ua{29.2} \\
        \texttt{OPT-350M}\textsubscript{\texttt{+ER}} 
        & $63.4$ & $\phantom{0}1.8$ & $66.2$
        & $91.3$ \ua{27.9} & $\phantom{00}0.1$ \db{1.8} & $91.4$ \ua{25.2} \\
        \texttt{OPT-1.3B}\textsubscript{\texttt{+ER}} 
        & $68.4$ & $\phantom{0}2.1$ & $70.3$
        & $91.3$ \ua{22.9} & $\phantom{00}0.1$ \db{2.0} & $91.4$ \ua{21.1} \\
        \texttt{OPT-2.7B}\textsubscript{\texttt{+ER}} 
        & $68.6$ & $\phantom{0}2.1$ & $70.4$
        & $91.4$ \ua{22.8} & $\phantom{00}0.1$ \db{2.0} & $91.5$ \ua{21.1} \\
        \texttt{OPT-6.7B}\textsubscript{\texttt{+ER}} 
        & $71.7$ & $\phantom{0}2.0$ & $73.4$
        & $91.4$ \ua{19.7} & $\phantom{00}0.0$ \db{2.0} & $91.4$ \ua{18.0} \\
        \texttt{OPT-13B}\textsubscript{\texttt{+ER}} 
        & $71.5$ & $\phantom{0}2.2$ & $73.7$
        & $91.4$ \ua{19.9} & $\phantom{00}0.1$ \db{2.1} & $91.4$ \ua{17.7} \\
        \bottomrule
        \end{tabular}
    }
    \caption{Performance of models on our continual GLUE setup (10 seeds). Performance gain is compared using correctness of the final tool response rather than the API call correctness. Notation is the same as in \autoref{tab:cl-arithmetic}. 
    }
    \label{tab:cl-glue}
\end{table*}

\begin{table*}[ht]
    \centering
    \resizebox{\linewidth}{!}{%
        \begin{tabular}{lccccccccc}
        \toprule
        \multicolumn{1}{c}{\textbf{\textsc{Model}}} & 
        \multicolumn{3}{c}{\textbf{\textsc{Without Tools}}} & 
        \multicolumn{3}{c}{\textbf{\textsc{With Tools (API Call)}}} &
        \multicolumn{3}{c}{\textbf{\textsc{With Tools (Final Answer)}}} \\
        & \textsc{Acc. ($\uparrow$)} & \textsc{For. ($\downarrow$)} & \textsc{L-A.} ($\uparrow$) 
        & \textsc{Acc. ($\uparrow$)} & \textsc{For. ($\downarrow$)} & {\textsc{L-A.} ($\uparrow$)} 
        & \textsc{Acc. ($\uparrow$)} & \textsc{For. ($\downarrow$)} & {\textsc{L-A.} ($\uparrow$)} \\
        \midrule
        \texttt{OPT-125M} 
        & $27.0_{\pm 1.1}$ & $92.4_{\pm 4.6}$ & $56.5_{\pm 2.9}$
        & $25.4_{\pm 1.5}$ & $100.0_{\pm 0.0}$ & $99.3_{\pm 0.0}$
        & $23.4_{\pm 1.4}$ & $100.0_{\pm 0.0}$ & $91.4_{\pm 0.1}$ \\
        \texttt{OPT-350B} 
        & $27.2_{\pm 0.9}$ & $94.4_{\pm 5.3}$ & $57.3_{\pm 2.9}$
        & $25.3_{\pm 1.6}$ & $100.0_{\pm 0.0}$ & $99.2_{\pm 0.0}$
        & $23.3_{\pm 1.4}$ & $100.0_{\pm 0.0}$ & $91.5_{\pm 0.0}$ \\
        \texttt{OPT-1.3B} 
        & $27.2_{\pm 1.4}$ & $89.7_{\pm 5.3}$ & $67.3_{\pm 2.9}$
        & $25.4_{\pm 1.3}$ & $\phantom{0}99.9_{\pm 0.0}$ & $99.2_{\pm 0.1}$
        & $23.4_{\pm 1.1}$ & $\phantom{0}99.7_{\pm 0.1}$ & $91.4_{\pm 0.0}$ \\
        \texttt{OPT-2.7B} 
        & $28.9_{\pm 1.8}$ & $93.2_{\pm 3.9}$ & $69.2_{\pm 3.6}$
        & $25.6_{\pm 1.6}$ & $\phantom{0}99.9_{\pm 0.1}$ & $99.0_{\pm 0.0}$
        & $23.5_{\pm 1.4}$ & $\phantom{0}99.4_{\pm 0.3}$ & $91.2_{\pm 0.1}$ \\
        \texttt{OPT-6.7B} 
        & $30.2_{\pm 1.2}$ & $92.7_{\pm 4.3}$ & $71.7_{\pm 3.5}$
        & $25.8_{\pm 1.4}$ & $\phantom{0}99.9_{\pm 0.1}$ & $99.3_{\pm 0.0}$
        & $23.6_{\pm 1.2}$ & $100.0_{\pm 0.0}$ & $91.4_{\pm 0.1}$ \\
        \texttt{OPT-13B} 
        & $29.9_{\pm 1.1}$ & $93.4_{\pm 3.4}$ & $72.1_{\pm 3.2}$
        & $25.7_{\pm 1.4}$ & $\phantom{0}99.9_{\pm 0.1}$ & $99.2_{\pm 0.0}$
        & $23.5_{\pm 1.2}$ & $100.0_{\pm 0.0}$ & $91.4_{\pm 0.1}$ \\
        \midrule
        \texttt{OPT-125M}\textsubscript{\texttt{+ER}} 
        & $60.7_{\pm 0.6}$ & $\phantom{0}1.9_{\pm 0.5}$ & $62.2_{\pm 0.8}$
        & $99.2_{\pm 0.1}$ & $\phantom{00}0.1_{\pm 0.1}$ & $99.3_{\pm 0.1}$
        & $91.3_{\pm 0.1}$ & $\phantom{00}0.1_{\pm 0.1}$ & $91.4_{\pm 0.1}$ \\
        \texttt{OPT-350M}\textsubscript{\texttt{+ER}} 
        & $63.4_{\pm 0.5}$ & $\phantom{0}1.8_{\pm 0.4}$ & $66.2_{\pm 0.8}$
        & $99.3_{\pm 0.1}$ & $\phantom{00}0.1_{\pm 0.1}$ & $99.4_{\pm 0.1}$
        & $91.3_{\pm 0.1}$ & $\phantom{00}0.1_{\pm 0.1}$ & $91.4_{\pm 0.1}$ \\
        \texttt{OPT-1.3B}\textsubscript{\texttt{+ER}} 
        & $68.4_{\pm 0.7}$ & $\phantom{0}2.1_{\pm 0.6}$ & $70.3_{\pm 0.9}$
        & $99.2_{\pm 0.1}$ & $\phantom{00}0.1_{\pm 0.1}$ & $99.3_{\pm 0.1}$
        & $91.3_{\pm 0.1}$ & $\phantom{00}0.1_{\pm 0.1}$ & $91.4_{\pm 0.1}$ \\
        \texttt{OPT-2.7B}\textsubscript{\texttt{+ER}} 
        & $68.6_{\pm 0.6}$ & $\phantom{0}2.1_{\pm 0.5}$ & $70.4_{\pm 1.1}$
        & $99.4_{\pm 0.1}$ & $\phantom{00}0.0_{\pm 0.1}$ & $99.5_{\pm 0.1}$
        & $91.4_{\pm 0.1}$ & $\phantom{00}0.1_{\pm 0.1}$ & $91.5_{\pm 0.1}$ \\
        \texttt{OPT-6.7B}\textsubscript{\texttt{+ER}} 
        & $71.7_{\pm 0.8}$ & $\phantom{0}2.0_{\pm 0.6}$ & $73.4_{\pm 0.9}$
        & $99.4_{\pm 0.1}$ & $\phantom{00}0.0_{\pm 0.1}$ & $99.4_{\pm 0.1}$
        & $91.4_{\pm 0.1}$ & $\phantom{00}0.0_{\pm 0.1}$ & $91.4_{\pm 0.1}$ \\
        \texttt{OPT-13B}\textsubscript{\texttt{+ER}} 
        & $71.5_{\pm 0.7}$ & $\phantom{0}2.2_{\pm 0.4}$ & $73.7_{\pm 1.0}$
        & $99.4_{\pm 0.1}$ & $\phantom{00}0.1_{\pm 0.1}$ & $99.5_{\pm 0.1}$
        & $91.4_{\pm 0.1}$ & $\phantom{00}0.1_{\pm 0.1}$ & $91.4_{\pm 0.1}$ \\
        \bottomrule
        \end{tabular}
    }
    \caption{Average accuracy, forgetting and learning accuracy for models on our continual GLUE setup (averaged across 10 seeds). Tool performance is separated into the ability to generate the correct API call as well as the ability of the tool to produce the correct response given the correct API call (incorrect API calls are treated as incorrect).}
    \label{tab:cl-glue-se}
\end{table*}

\subsection{Additional Plots}
\begin{figure}[ht]
    \centering
    \includegraphics[width=\linewidth]{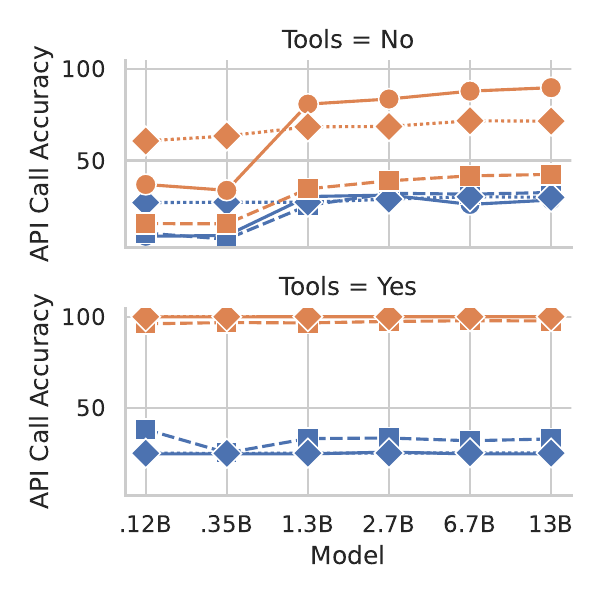}
    \caption{Accuracy of API calls on all tasks.}
    \label{fig:Accuracy-API-all-tasks}
\end{figure}
\begin{figure}[ht]
    \centering
    \includegraphics[width=\linewidth]{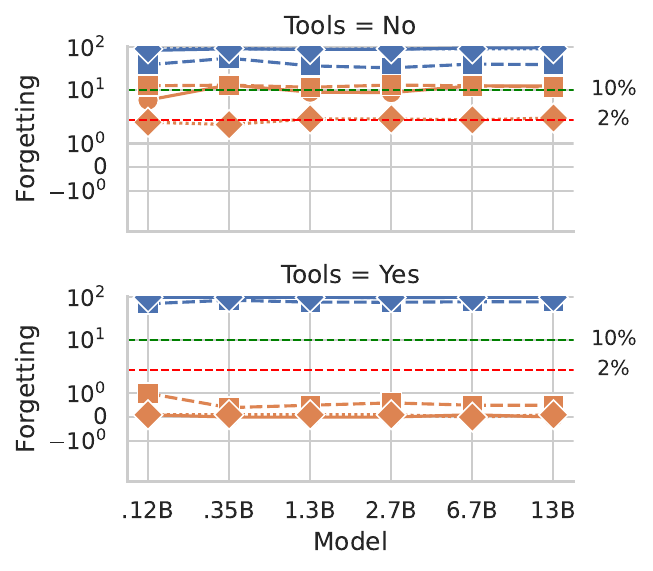}
    \caption{\autoref{fig:Forgetting-all-tasks} with forgetting values on the log-scale. Horizontal lines indicate $2$\% and $10$\% forgetting rates.}
    \label{fig:Forgetting-log-all-tasks}
\end{figure}

\end{document}